\documentclass{article}

\usepackage[final]{ml4eng_2020}

\usepackage[utf8]{inputenc} 
\usepackage[T1]{fontenc}    
\usepackage{hyperref}       
\usepackage{url}            
\usepackage{booktabs}       
\usepackage{amsfonts}       
\usepackage{nicefrac}       
\usepackage{microtype}      
\usepackage{graphicx} 
\usepackage{algorithm}
\usepackage[noend]{algpseudocode}
\usepackage{mathtools}
\usepackage{xcolor}
\usepackage{multirow}
\usepackage{multicol}

\title{Rethink AI-based Power Grid Control: Diving Into Algorithm Design}

\newcommand*{\affaddr}[1]{#1} 
\newcommand*{\affmark}[1][*]{\textsuperscript{#1}}
\newcommand*{\email}[1]{\texttt{#1}}

\author{%
Xiren Zhou\affmark[1], Siqi Wang\affmark[2], Ruisheng Diao\affmark[2], Desong Bian\affmark[2], Jiajun Duan\affmark[2] and Di Shi\affmark[2]\\
\affaddr{\affmark[1]Columbia University}\\
\email{\affmark[1] xz2754@columbia.edu} \\
\affaddr{\affmark[2]Global Energy Interconnection Research Institute North America(GEIRINA) }\\
\email{\affmark[2]\{siqi.wang, ruisheng.diao, desong.bian, jiajun.duan, di.shi\}@geirina.net }\\
}

\begin{document}

\maketitle
\begin{abstract}
Recently, deep reinforcement learning (DRL)-based approach has shown promise in solving complex decision and control problems in power engineering domain. In this paper, we present an in-depth analysis of DRL-based voltage control from aspects of algorithm selection, state space representation, and reward engineering. To resolve observed issues, we propose a novel imitation learning-based approach to directly map power grid operating points to effective actions without any interim reinforcement learning process. The performance results demonstrate that the proposed approach has strong generalization ability with much less training time. The agent trained by imitation learning is effective and robust to solve voltage control problem and outperforms the former RL agents. 
\end{abstract}

\section{Introduction}
Nowadays, the rapid development of artificial intelligence (AI) technologies provides new ideas and solutions for solving many challenges in the field of power grid operation and control. The application of deep reinforcement learning has been extensively explored to solve complex power engineering problems, such as grid emergency control [1], real-time autonomous energy management [2] and topology adjustment [3]. Specifically, DRL-based power grid control paradigm has become a hot spot and provides effective future development direction for both power system research and engineering community. The pioneer work in [4] presents a novel autonomous control paradigm called "Grid Mind" to derive fast and effective controls in real time with Deep Q Network (DQN) agent to eliminate voltage violations. Later on, the follow-up work in [5] expands the findings of [4] and modified the DQN algorithm to improve control performance by avoiding the choice of same actions multiple times and normalizing the observations. Similar ideas have been further expanded in [6] and [7] for controlling the voltage setpoint of generators and PV-converters in a continuous manner by employing Deep Deterministic Policy Gradient (DDPG) agent. However, the previous work is mainly focused on demonstrating that the state-of-art DRL algorithm can be plugged in to the power grid control framework. Different from the previous exploration, in this paper, we investigate the underlying nature of power grid voltage control and revisit it from the Markov Decision Process (MDP) point of view, mainly focusing on tuning the design of the problem formulation, i.e., the choice of state representation, the DRL algorithm selection, and the reward engineering. Moreover, we present the lessons learnt from training an effective DRL agent using the real world power grid data collected from the control center of SGCC Jiangsu Electric Power Company. We further propose an imitation learning-based approach to resolve the voltage control problem based upon the findings of the training process and demonstrate that the imitation learning approach is also an effective alternative without any reinforcement learning component for this type of voltage control problem. 

\section{Problem formulation}
The main objective of power system voltage control is to maintain bus voltage profiles within the predefined bounds and at the same time keeping transmission line flows within limits. The state is the current operating condition of the power grid represented by the power flow snapshot at a given timestamp. We model the power grid control problem as an MDP as follows: 
\begin{itemize}
    \item $S$: an infinite state space of continuous-valued state representation. $S \in \mathbb{R}^{\textup{n}_s} $.
    \\Three types of measurement values can be adopted to construct state space: bus values (bus voltage $V_{m}$ and bus angle $V_{a}$), branch values (line flow $S_{line}$), and generator values (active power $P_{g}$ and reactive power $Q_{g}$). 
    \item $A$: an infinite action space of continuous-valued action vector. $A \in \mathbb{R}^{\textup{n}_a} $. 
    \\The voltage setpoints of generators within each power plant are used to control bus voltages. Therefore n$_{a}$ represents the number of power plants in the studied power grid. The plant voltage value is bounded within the range of [0.9, 1.1] in p.u.
    \item $P$: a transition dynamics model that specifies $P(s'|s, a)$.
    \\In essence, the transition probability $P(s'|s, a)$ is determined and dominated by the physics laws and all configurations of its underlying environment. 
    In this design, we use an in-house power flow solver (simulator) to model the transitions from state $s$ to state $s'$ after applying action $a$, i.e., $P(s'|s, a)=1$ where $s'$ is obtained from the power flow solver given the input $s$ and $a$. Thus, $P$ is a sparse/one-hot matrix (of infinite dimension) in our MDP. 
    \item $R$: a reward function that maps a state action pair to a real number. 
    \\Due to the sparse property of $P$ mentioned above, the reward at time step $t$ can also be determined by the resulting state at time step $t+1$. i.e., $r_t=R(s_t, a_t)=R(s_{t+1})$. 
    \item $\gamma$: the discount factor. It is fixed to 0.99 throughout the paper.
\end{itemize}
Moreover, we define a set of terminal states $T$ which is composed of all the states without voltage violation for all buses and without line flow violation for all lines. A state $s$ is "unsuccessful" iff $s \notin T$, meaning that there exists either voltage violations or line flow violations at the current operating condition. Likewise, we define any $s \in T$ as a "successful" state. The MDP is episodic by defining terminal states as above and enforcing a horizon limit for each episode (e.g., 50, 100, 1000, etc.). 

In summary, the ultimate goal for the DRL agent is to find a policy $\pi$ (a mapping from state to action) that can eliminate both voltage violations and line flow violations as quickly as possible. In other words, given an unsuccessful initial state, the policy is desired to be able to make effective sequence of decisions that leads to a successful state in as few number of steps as possible. 
\section{Data preparation}
We collected 10,433 power flow snapshots from the control center of SGCC Jiangsu Power Company, which represents the power grid operating conditions from January to March, 2020. The snapshots are stored in *.dat format, each of which can be treated as an unsuccessful initial state $s_0$. To eliminate the seasonal impact of co-relation, the entire dataset is randomly split into 9,433 and 1,000 for training and testing, respectively, i.e.:
\begin{equation}
    S_{\textup{train}}=\{s_0^{(1)}, s_0^{(2)}, ..., s_0^{(9433)}\}; 
    S_{\textup{test}}=\{s_0^{(9434)}, s_0^{(9435)}, ..., s_0^{(10433)}\}
\end{equation}

\section{RL-based power grid voltage control}
\subsection{Algorithm selection}
Both of value based RL algorithm (i.e., DQN) and policy gradient based RL algorithm (i.e., DDPG) have been studied in previous work to some extent. But policy gradient RL algorithms tend to be more appropriate and effective in solving power grid control problem for the following reasons: 1) policy gradient based algorithms are proven to converge to at least local optimum (under certain conditions) whereas DQN does not have such guarantee mathematically [8]; 2) an ordinary DQN is difficult to deal with infinite continuous action space while both action space and state space of voltage control are continuous. In this work, we consider soft actor critic (SAC) as our policy gradient RL algorithm instead of DDPG to enable more incentives for randomness in actions, which is important in finding a good plant voltage point through exploration [9].

\subsection{Reward design strategies}
As mentioned before, the goal of training RL agents is to move from an unsuccessful state $s_0$ to a successful state in as few steps as possible. However, in SAC formulation, the agent is optimized directly upon the discounted sum of rewards instead of the number of steps per episode. Therefore, a well-designed reward function plays a critical role in training an effective agent. The agent must have the capability to learn lessons efficiently from the variation of the rewards to shorten its number of steps within an episode.





Intuitively, the essence of the policy gradient formula is to increase the probability of taking the actions that receive large rewards and decrease the probability of taking the actions that receive small rewards. A good reward function design should be able to direct the policy network effectively through rewards and penalties (i.e. small positive rewards or even negative rewards). With this assumption, we define our reward function R by dividing it up into two separate functions $R_{-}$ and $R_{+}$, according to the types of a transition step (either successful or unsuccessful). i.e.,
\begin{equation}
r_t =R(s_t, a_t)=
\begin{cases}
 &R_{-}(s_{t+1}) \text{, if } s_{t+1} \notin T \\ 
 &R_{+}(s_{t+1}) \text{, if } s_{t+1} \in T
\end{cases} \label{eq_2}
\end{equation}

And we propose the following two reward design strategies:
\begin{enumerate}
    \item $R_{-}(s)= f_{\textup{penalty}}(V_m, S_{line})$; \label{strategy:1}\\
    $R_+(s)$ is a fixed non-negative constant.
    \item CartPole-style reward: $R \equiv -1$. \label{strategy:2}

\end{enumerate}
where $f_{\textup{penalty}}=\alpha \sum_{i} \textup{line\_overflow}[i] + \beta \sum_{j}\textup{bus\_violation}[j]$ computes the penalty (negative reward) of a given state according to its $S_{line}$ and $V_m$.

The overflow of the $i$th line and the voltage violation of the $j$th bus are defined as following:
\begin{equation}
    \begin{aligned}
    &\textup{line\_overflow}[i]=\max\{S_{line}[i]-\textup{line\_limit}[i], 0\}^2 \\
    &\textup{bus\_violation}[j]=\max\{(V_m[j]-\textup{bus\_lower\_limit}[j])(V_m[j]-\textup{bus\_upper\_limit}[j]), 0\}
    \end{aligned}
    \label{eq: penelty function}
\end{equation}
$\alpha$ and $\beta$ are used to balance the importance between line overflow and bus voltage violation, which are set to $-0.1$ and $-1000$ respectively throughout this paper.

Since the optimization objective is the average number of steps needed to solve an unsuccessful state, which is basically the length of the episode, we attempt to make the agent learn directly from it. Hence, strategy \ref{strategy:2} is proposed to make the length of an episode reflected by the return value. We call it "CartPole-style" reward since this is similar to the reward style of the CartPole environment [10]. 

It is the goal that under either of the two strategies above, the agent will converge to a theoretically optimal policy that solves all voltage violation cases in one step. However, it is difficult to achieve in real world, because power grid is such a complex system that we are not able to theoretically ensure it is Markovian. Meanwhile, for some voltage violation cases, there might not exist any one-step solution, given that we only enable control over plant voltage value settings.

In the next section, we demonstrate the experiment results of different reward function designs, where we end up with a best reward design pattern and obtain some insights into power grid MDP. 

\subsection{Experiments \& analysis}
As shown in Table \ref{tab:strategy 1 result}, we tried different values for $R_+$ for reward design strategy \ref{strategy:1}. A higher $R_+$ basically makes the training more efficient and converge to an optimal policy more quickly. The SAC algorithm with $R_+=0$ fails to find an optimal policy.
\label{section: reward design analysis}
\begin{table}[H]
\caption{Training time needed til finding the optimal policy under different $R_+$ for strategy \ref{strategy:1}}
    \centering
    \begin{tabular}{c|ccccc}
    $R_+$ & 0 & 1 & 10$\sim$20 & 50 & 80$\sim$1000 \\
    \hline
    training steps & fail to converge & 3.4k & 1.3k & 1k & 0.9k
    \end{tabular}
    \label{tab:strategy 1 result}
\end{table}
\vspace{-10pt}
 For strategy \ref{strategy:2}, with $R \equiv -1$, the length of an episode is directly reflected by its sum of reward. However, it fails to converge to an optimal policy. We then adjusted $R_+$ to 0 or 1, which is different from the original "CartPole-style" setting, since it is inappropriate to give the agent a penalty even on a successful step. Unfortunately, the training diverges as well. We finally set a large value (1000) for $R_+$, which eventually works well. 
\begin{table}[H]
\caption{Training time needed til finding the optimal policy under different $R_+$ for strategy \ref{strategy:2}}
    \centering
    \begin{tabular}{c|cccc}
    $R_+$ & -1 & 0 & 1 & 1000 \\
    \hline
    training steps & fail to converge & fail to converge & fail to converge & 0.9k
    \end{tabular}
    \label{tab:strategy 2 result}
\end{table}
\vspace{-10pt}

From these experiments, we find out that a higher positive reward makes the agent learn lessons more quickly and efficiently. As for the negative reward on any unsuccessful step, according to formula \ref{eq: penelty function}, the penalty increases quadratically as the line flow or bus voltage goes beyond the limit, which is a common design in this field. However, It turns out that a well-designed meaningful $R_-$ does not play a such significant role as $R_+$; even with $R_-$ as a constant, as long as $R_+$ is large enough, the agents can learn very well. 

These observations imply that the agent learns significantly from the last successful steps. Those unsuccessful steps give little helpful information to the RL agent. It inspires us to train the agent with only successful steps and throw away all unsuccessful steps, which leads to the imitation learning based method that will be covered in the next section.

\section{Imitation learning-based power grid control}

\begin{algorithm}[!htbp]
\caption{Imitation Learning for Training a Power Grid Control Agent} 
\label{algorithm:IL}
\begin{multicols}{2}
\begin{algorithmic}
\State Initialize: policy network $\pi$ with random weights $\theta$
\State $D$ = COLLECT SUCCESSFUL STEPS($S_{\textup{train}}$)
\State $l_{\textup{objective}}=1$ \# set an objective average episode's length to terminate the runtime.
\While{True}
\State Train $\theta$ with an optimizer (eg., Adam) on dataset $D$ for an epoch
\State $l$ = EVALUATE EPISODE LENGTH($S_{\textup{test}}$, $\pi_{\theta}$)
\If {$l \leq l_{\textup{objective}}$}
\State \textbf{terminate}
\EndIf
\EndWhile
\\
\Procedure{Collect Successful Steps}{$S$}
\State Initialize $D=\emptyset$
\State Set policy $\pi_{\textup{collect}}$ to be an arbitrary policy (eg., random policy, trained SAC policy, etc.)
\State $n= $ number of successful steps to collect (eg., 10000)
\State $t_{\textup{limit}}=1000$ \# set a horizon limit
\While{$|D|<n$}
\State Randomly sample a state $s_0$ from $S$
\For {$t=0,1,2...,t_{\textup{limit}}-1$}
\State $a_t=\pi_{\textup{collect}}(s_t)$
\State $s_{t+1}$ $\leftarrow$ perform $a_t$ on $s_t$
\If{$s_{t+1} \in T$}
\State $D=D \cup \{(s_t, a_t)\}$
\State break
\EndIf
\EndFor
\EndWhile
\State \Return $D$
\EndProcedure
\\
\Procedure{Evaluate Episode Length}{$S$, $\pi$}
\State Initialize $L=\emptyset$
\State $n_{\textup{episodes}}=50$ \# set number of episodes(cases) to evaluate
\State $t_{\textup{limit}}=50$ \# set a horizon limit
\For{$i=1,2,...,n_{\textup{episodes}}$}
\State Randomly sample a state $s_0$ from $S$
\For {$t=0,1,2...,t_{\textup{limit}}-1$}
\State $a_t=\pi(s_t)$
\State $s_{t+1}$ $\leftarrow$ perform $a_t$ on $s_t$
\If{$s_{t+1} \in T$ \textbf{or} $t+1=t_{\textup{limit}}$}
\State $L=L \cup \{t+1\}$
\State break
\EndIf
\EndFor
\EndFor
\State \Return $\mathbb{E}_{l \in L}[l]$
\EndProcedure
\end{algorithmic}
\end{multicols}
\end{algorithm}
Inspired by the study in Section \ref{section: reward design analysis}, an imitation learning method is proposed to train an agent with only successful steps. Specifically, the proposed method does not incorporate any reinforcement learning component except a policy neural network mapping from a state vector to an action vector. The network is trained by supervised learning on a dataset $D$:
\begin{equation}
D=\{(s_t, a_t)|\textup{where } s_{t+1} \in T\}
\end{equation}

\section{Performance \& analysis}
\subsection{Random policy baseline performance}
For each $s_0 \in S_{\textup{train}} \cup S_{\textup{test}}$, a random agent is employed to interact with the environment. The simulation is conducted multiple times (episodes) for each $s_0$ and the average number of steps needed to solve each case are recorded. The maximum episode length is 1000. If a case can never be solved within 1000 steps, it is considered as an "unsolvable" case. Figure \ref{fig:PolicyPerformanceComparison}(a) shows the random policy performance on the entire dataset of 10,433 cases. There are totally 142/10433 (1.36\%) unsolvable cases.  

\subsection{SAC agent performance}
\begin{figure}[!htbp]
  \centering
  \includegraphics[scale=0.3]{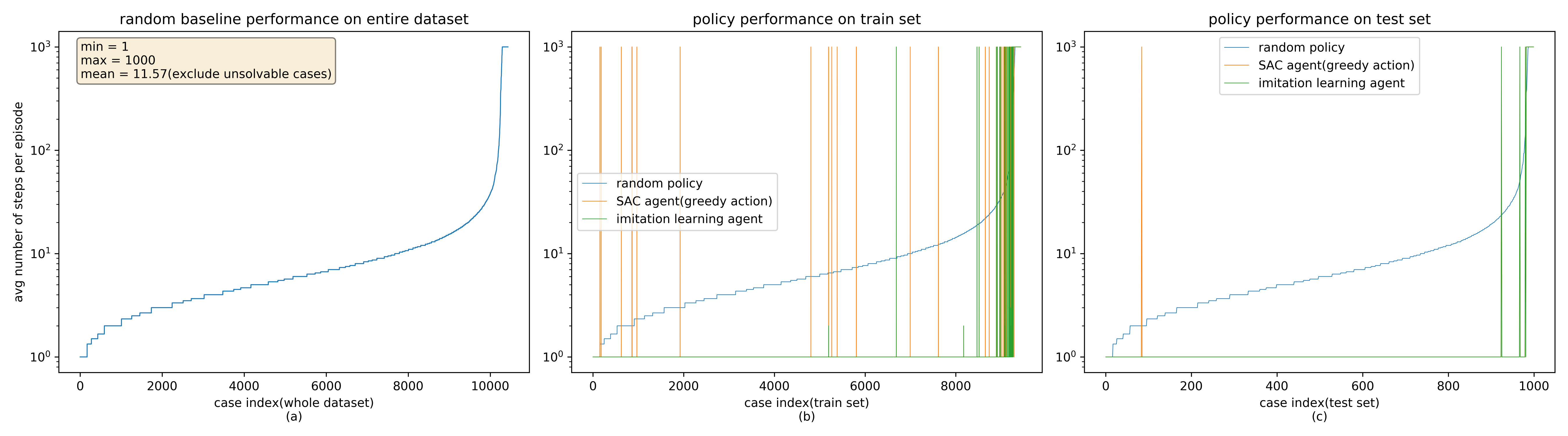}
  \caption{Policy performance comparison. (Case indices are sorted according to random policy performance. A case is considered unsolvable if its y value reaches $10^3$.)}
  \vspace{-10pt}
  \label{fig:PolicyPerformanceComparison} 
\end{figure}

The SAC agent was trained with $R_+=1000$. The evaluation result of the agent is shown in Figure \ref{fig:PolicyPerformanceComparison} and Table \ref{PolicyPerformance}. We evaulated the agent's policy with its stochastic (i.e., the action is sampled from a normal distribution given its learned mean and standard deviation) and greedy (i.e., the action is solely determined by its mean, which means a deterministic policy) variants respectively. The greedy SAC policy is able to solve all cases in only one step; but there exist more unsolvable cases, compared with random policy. The stochastic SAC policy, on the other hand, has a higher solvable rate (98.5\% on train set and 98.4\% on test set) but needs more steps on average to solve a voltage violation case. One reason is that the state representation is still possibly not a Markovian representation (considering the complexity of power grid system), and adding randomness can alleviate the partial observability problem [11]. This is why a policy with distributed action or a random policy is able to solve more cases (regardless of steps cost) than a deterministic policy. Despite the advantage of stochasticity, the deterministic policy is superior in that it solves a case instantly (one step). In practice, it relies on the grid operators' engineering judgement to decide which policy is better. 

\subsection{Imitation learning agent performance}
A dataset $D$ is collected according to Algorithm \ref{algorithm:IL} by running random policy, which contains $\sim$20k successful transitions corresponding to $S_{\textup{train}}$. In other words, any of the successful transition is generated within an episode starting with a certain initial state in $S_{\textup{train}}$. Note that there might be multiple transitions in $D$ corresponding to the same $s_0$ in $S_{\textup{train}}$. The network architecture is the same as the policy network of the former SAC agent (see \hyperref[appendix]{Appendix}). The training takes only three epochs until finding the optimal policy, which is much faster than the SAC algorithm. As shown in Figure \ref{fig:PolicyPerformanceComparison} and Table \ref{PolicyPerformance}, the imitation learning agent performs even better than the SAC agent with a higher solvable rate. It can solve any solvable case in one step. The strong generalization ability of the imitation learning agent can be viewed in two aspects: 1) $D$ was collected by exploring cases in $S_{\textup{train}}$, and the agent is able to solve most cases in $S_{\textup{test}}$ in one step; 2) consider $(s, a) \in D$ which is the successful step of an episode starting with $s_0^{(i)} \in S_{\textup{test}}$, since $D$ was collected by running a random policy, mostly $s \neq s_0^{(i)}$ (the case is not solved within one step.) E.g., say $s=s_5^{(i)}$, training with $(s_5^{(i)}, a_5^{(i)})$, the agent is able to learn an $a_0^{(i)}$ which instantly solves this case in one step. 
\subsection{PCA analysis of the state space}
As demonstrated in Figure \ref{fig:PolicyPerformanceComparison}, and in Table \ref{PolicyPerformance}, there are a small number of extreme cases that are unsolvable no matter what policy it is. In order to have a deep understanding about the states of the environment, the principle component analysis (PCA)is conducted to investigate the cluster distribution of unsolvable cases versus solvable cases in Euclidean space.
Specifically, for $\forall s \notin T$, a random agent is initiated with allowable maximum steps as 1000 for multiple times. If $s$ can never be solved within 1000 steps during multiple trials, we consider $s$ as an "unsolvable" case (state); otherwise, it's a "solvable" case (state).

\begin{figure}[!htbp]
 \centering
 \includegraphics[width=0.65\textwidth,height=0.3\textwidth]{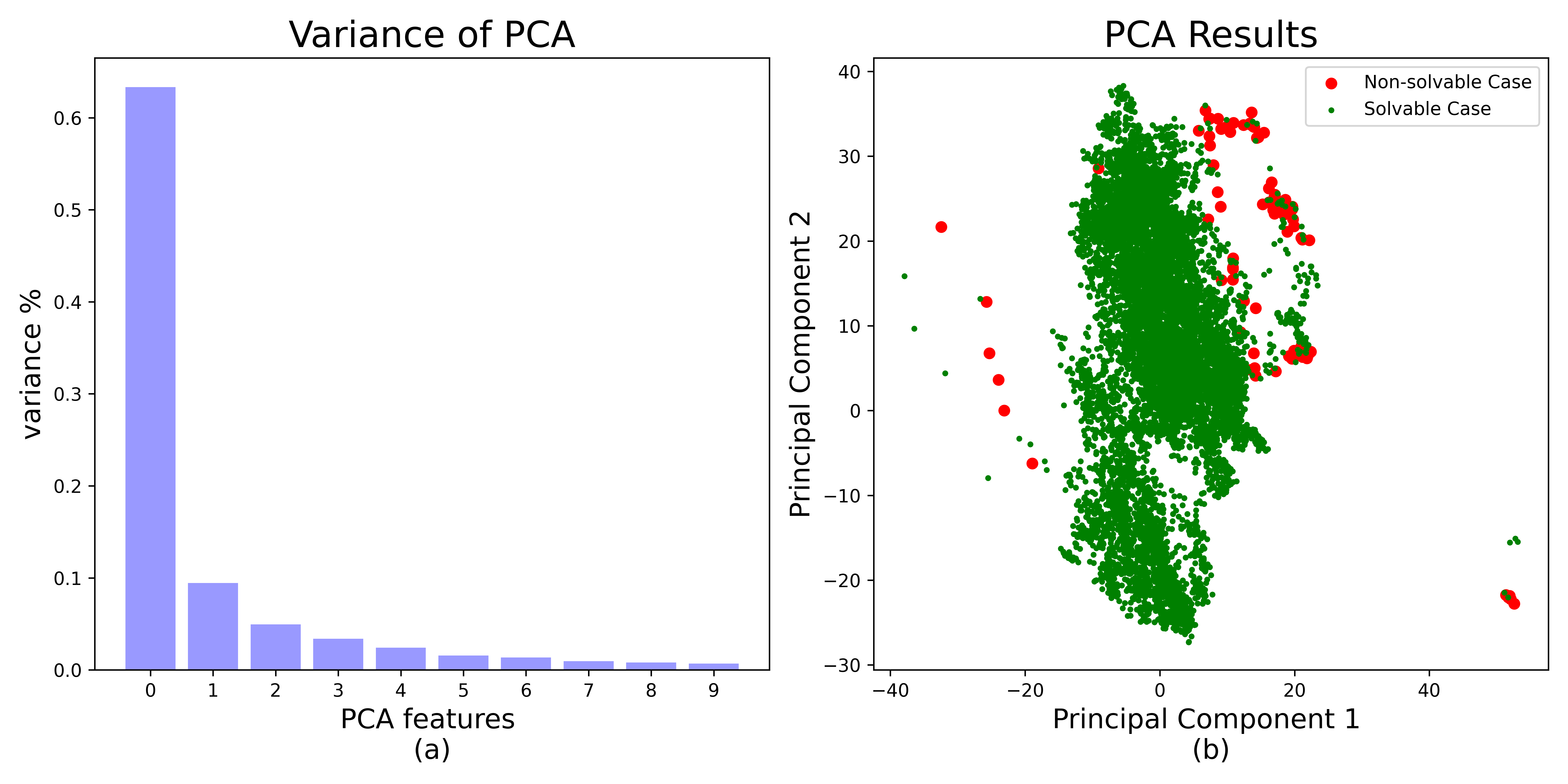}
 \caption{PCA results: (a) Variations (b) Non-solvable v.s Solvable case}
 \vspace{-10pt}
 \label{fig: PCA} 
\end{figure}

The full state has a dimension of 1,442. Figure \ref{fig: PCA}(a) verifies that the first two components can explain the majority of the variance in the data and Figure \ref{fig: PCA}(b) shows there is a clear boundary between unsolvable cases and solvable cases. This fact corresponds to the fact in complex power grid that there exists a small amount of extreme cases which can be proven unsolvable in the case of adjusting plant voltage setpoint only.

\begin{table}
  \caption{Policy performance details}
  \label{PolicyPerformance}
  \centering
  \begin{tabular}{lll}
    \toprule
    Policy  &  Number of unsolvable cases  &  Avg steps to solve a case \\
    \midrule
    \multirow{2}*{random policy} & \textbf{train set: 127/9433(1.35\%)}   & train set: 11.53 \\
    & \textbf{test set: 13/1000(1.3\%)}   & test set: 11.98 \\
    \cline{2-3}
    \multirow{2}*{SAC agent(normal distributed action)} & train set: 138/9433(1.46\%)   & train set: 3.17 \\
    & test set: 16/1000(1.6\%)   & test set: 2.91 \\
    \cline{2-3}
    \multirow{2}*{SAC agent(greedy action)} & train set: 226/9433(2.40\%)   & \textbf{train set: 1} \\
    & test set: 22/1000(2.2\%)   & \textbf{test set: 1} \\
    \cline{2-3}
    \multirow{2}*{imitation learning agent} & train set: 196/9433(2.08\%)   & \textbf{train set: 1} \\
    & test set: 21/1000(2.1\%)  & \textbf{test set: 1} \\
    \bottomrule
  \end{tabular}
\end{table}

\section{Conclusion}
In this work, we revisited the previous DRL-based voltage control problem of power grid. We performed an in-depth analysis on algorithm selection, state space representation, and reward engineering. Based upon the analysis result, we realize that the agent mostly learn lessons from the positive rewards of the last successful steps. Thus, we optimize the reward design which results in a sample-efficient SAC-based approach that converges to the optimal policy very quickly. Furthermore, we proposed a novel imitation learning-based approach to perform power grid voltage control. The training and testing results show that the trained imitation learning agent has strong generalization ability which even outperforms the RL agent with the same policy network architecture. Meanwhile, the imitation learning based method does not involve any complex hyper-parameter tuning or design of a reward function, and requires less training time to converge to the optimal policy.


\section*{Broader Impact}
The rapid development of renewable energy brings more and more complexity to the control of power grid. The traditional control methods based on operator's experience are difficult to cope with the complex and changeable power grid operating conditions (especially under unknown operating conditions). How to effectively realize rapid regulation of power grid is an urgent problem to be solved. With the fast development of AI technology, it is promising to provide operators with accurate and timely control plans, to improve the control efficiency during incidents. This work provides an in-depth analysis on the DRL-based methodologies for autonomous voltage control for power grid operation. Key aspects of MDP are thoroughly investigated for improving the performance of RL agents. During this process, we found that such MDP formulation aims at obtaining one-step control to fix voltage violations once detected; then we propose an imitation learning-based approach to achieve this goal, the effectiveness of which is verified via massive simulation studies conducted on actual power grid operating conditions. It is our hope that this work can help research community in better understanding the underlining principles of such control problems and promote AI-based solutions towards real-world implementation.

\section*{References}
\small
[1] Q. Huang, R. Huang, W. Hao, J. Tan, R. Fan, and Z. Huang, “Adaptive power system emergency control using deep reinforcement learning,” \textit{IEEE Trans. Smart Grid}, vol. 11, no. 22, pp. 1171-1182, 2020.

[2] Y. Ye, D. Qiu, X. Wu, G. Strbac and J. Ward, "Model-Free real-time autonomous control for a residential multi-energy system using deep reinforcement learning," \textit{IEEE Trans. Smart Grid}, vol. 11, no. 4, pp. 3068-3082, July 2020.

[3]	T. Lan, J. Duan, B. Zhang, D. Shi, Z. Wang, R. Diao, and X. Zhang, “AI-based autonomous line flow control via topology adjustment for maximizing time-series ATCs,” \textit{https://arxiv.org/abs/1911.04263}, 2019.

[4] R. Diao, Z. Wang, D. Shi, etc., “Autonomous voltage control for grid operation using deep reinforcement learning,” \textit{IEEE PES General Meeting}, Atlanta, GA, USA, 2019.

[5]	B. L. Thayer and T. J. Overbye, “Deep reinforcement learning for electric transmission voltage control,” \textit{arXiv preprint arXiv:2006.06728}, 2020.

[6]	J. Duan et al., "Deep reinforcement learning-based autonomous voltage control for power grid operations," \textit{IEEE Trans. Power Systems}, vol. 35, no. 1, pp. 814-817, Jan. 2020.

[7] C. Li, C. Jin, and R. K. Sharma, “Coordination of pv smart inverters using deep reinforcement learning for grid voltage regulation,” \textit{2019 18th IEEE International Conference On Machine Learning And Applications (ICMLA)}, pp. 1930–1937, 2019

[8] R. S. Sutton, D. Mcallester, S. Singh, and Y. Mansour. "Policy gradient methods for reinforcement learning with function approximation." In \textit{NIPS}, pages 1057–1063. MIT Press, 2000.

[9] T. Haarnoja, A. Zhou, P. Abbeel, and S. Levine. Soft actor-critic: Off-policy maximum entropy deep reinforcement learning with a stochastic actor. In \textit{International Conference on Machine Learning (ICML)}, volume 80, pages 1856–1865, 2018.

[10] G. Brockman et al., "OpenAI Gym", \textit{arXiv:1606.01540}, 2016

[11] Singh, Satinder, T. Jaakkola and Michael I. Jordan. “Learning without state-estimation in partially observable markovian decision processes.”  \textit{Proc. ICML-94}, pp. 284-292.

\section*{Appendix: }
\label{appendix}
All code implementations are based on \href{https://www.tensorflow.org/}{TensorFlow 2.3.0}. The SAC implementation is based on \href{https://www.tensorflow.org/agents}{TF-Agents 0.6.0}. The hyperparameters used for training are shown in Table \ref{sac-hyperparameter-table} and \ref{IL-hyperparameter-table}. 

\begin{table}[ht]
  \caption{Variables and physical quantities}
  \label{power_parameter_table}
  \centering
  \begin{tabular}{lll}
    \toprule
    Name             & Description                & Unit or value\\
    \midrule
    $V_{m}$          & voltage magnitude          & p.u   \\ 
    $V_{a}$          & voltage angle              & p.u   \\
    $S_{line}$         & line flow apparent power   & p.u    \\
    $P_{g}$  & active power of generator & MW \\
    $Q_{g}$  & reactive power of generator & MVar \\
    $\textup{n}_s$ & state dimension & 1442\\
    $\textup{n}_a$ & action dimension(number of plants) & 16\\
    \bottomrule
  \end{tabular}
\end{table}

\begin{table}[H]
  \caption{Hyperparameters used for training SAC agents}
  \label{sac-hyperparameter-table}
  \centering
  \begin{tabular}{lll}
    \toprule
    Hyperparameter     & Description     & Value \\
    \midrule
    critic\_obs\_fc & fully connected layers for observation in critic network & [512, 512]\\
    critic\_action\_fc & fully connected layers for action in critic network & [256]\\
    critic\_joint\_fc & fully connected layers after merging observations and actions & [256, 256]\\
    actor\_fc & fully connected layers for policy network & [512, 512]\\
    batch\_size & batch size  & 256     \\
    episode\_max\_len    & horizon limit during training & 50      \\
    lr    & learning rate   &  0.0003 \\
    $\tau_{\textup{target}}$ & Factor for soft update of the target networks & 0.005\\
    target\_update\_period & Period for soft update of the target networks & 1\\
    initial $\log(\alpha)$    & initial value of $\log(\alpha)$   &  1 \\
    collect\_episodes & number of episodes to collect per training step & 1\\
    \bottomrule
  \end{tabular}
\end{table}

\begin{table}[H]
  \caption{Hyperparameters used for imitation learning}
  \label{IL-hyperparameter-table}
  \centering
  \begin{tabular}{lll}
    \toprule
    Hyperparameter     & Description     & Value \\
    \midrule
    hidden\_fc\_layers & fully connected layers & [512, 512]\\
    batch\_size & batch size  & 32$\sim$512     \\
    lr    & learning rate   &  0.001 \\
    \bottomrule
  \end{tabular}
\end{table}

\end{document}